# Modelling Office Energy Consumption: An Agent Based Approach

Tao Zhang[1], Peer-Olaf Siebers[1], Uwe Aickelin[1]

[1] Intelligent Modelling & Analysis Group
School of Computer Science, University of Nottingham
Tao.Zhang@nottingham.ac.uk

**Abstract.** In this paper, we develop an agent-based model which integrates four important elements, i.e. organisational energy management policies/regulations, energy management technologies, electric appliances and equipment, and human behaviour, based on a case study, to simulate the energy consumption in office buildings. With the model, we test the effectiveness of different energy management strategies, and solve practical office energy consumption problems. This paper theoretically contributes to an integration of four elements involved in the complex organisational issue of office energy consumption, and practically contributes to an application of agent-based approach for office building energy consumption study.

## 1. Introduction

In the UK and many other industrialised countries, offices, as the basic units for buildings, are intensively distributed in big cities and urban areas. As climate change becomes a very important global political issue, the UK government has set a target of cutting $CO_2$ emission by 34% of 1990 levels by 2020. In the UK, the energy consumed in service sector took up 14% of overall energy consumption of the whole country in 2001. Most of the energy for service sector is used in various kinds of offices for heating, lighting, computing, catering and hot water. Thus, energy consumption in offices is one of the research areas which have significant importance for meeting the UK government's 2020 target.

Practically, the energy consumption in a modern office building is a very complex organisational issue involving four important elements: energy management policies/regulations made by the energy management division of an organisation, energy management technologies installed in the office building (e.g. metering, monitoring, and automation of switch-on/off technologies), types and numbers of the electric equipment and appliances in the office building (e.g. lights, computers and heaters), and more importantly staff's behaviour of using electric equipment and appliances in the office building. The four elements interact (Figure 1): the energy management division makes energy management policies/regulations based on the energy management technologies installed in the building; energy management technologies monitor and control the energy consumed by electric equipment and appliances, and also influence the behaviour of energy users in the building; energy users' behaviour of using electric equipment and appliances directly cause energy consumption. Yet the literature appearing in office building energy consumption has three lines of research. One line primarily focuses on building energy consumption prediction. Typical studies appearing in this area is the application of artificial neural networks (ANN) in building energy consumption (e.g. Anstett & Kreider, 1993; Olofsson et al, 1998; Karatasou et al, 2005; Gonzalez & Zamarreno, 2005). A second line of research primarily focuses on the energy management technologies in commercial buildings. Typical studies include building performance measurement (e.g. Zimmermann, 2007) and intelligent energy management systems (e.g. Hagras et al, 2003). A third line of research focuses on building energy consumption benchmarks, which are derived based on surveys of a large number of occupied building and include all energy uses. Building energy consumption benchmarks provide representative values for common building types, against which building energy managers can compare their buildings' actual energy performance (Carbon Trust, 2009). The three lines of research target parts of the office energy consumption issue, with ignorance of organisational energy management policies/regulations and human factors (i.e. staff behaviour) which are very important elements in office energy consumption. Motivated by a desire to comprehensively understand the complex organisational issue of office energy consumption, the first objective of the research reported in this paper is to provide a dynamic computational simulation model which integrates the aforementioned four interactive elements involved in office energy consumption.

Each organisation faces a dilemma in terms of energy consumption. On the one hand, it has to consume energy to satisfactorily meet the energy needs of its staff members and maintain typical comfort standards in its office buildings. One the other hand, it has to minimize its energy consumption through effective organisational energy management policies/regulations, in order to reduce its energy bills. A second objective of the research reported in this paper is to develop a multi-agent decision-making framework to help organisations make proper energy management policies/regulations to deal with this dilemma: maintaining efficient energy consumption to satisfactorily meet staff members' energy needs, whilst minimizing the energy consumption within the whole organisation, without significant investments on or changes to the current energy management technologies.

The paper is structured as follows. In the second section, we use a mathematical model to theoretically explain the energy consumption in office buildings. In the third section, we present our agent-based model of office energy consumption based on a case study in an academic building in Jubilee Campus, University of Nottingham. In the fourth section, we analyze the outputs from the simulation, and draw some energy management strategy implications. In the fifth section, we discuss the model, and in the sixth section, we summarize and conclude the study.

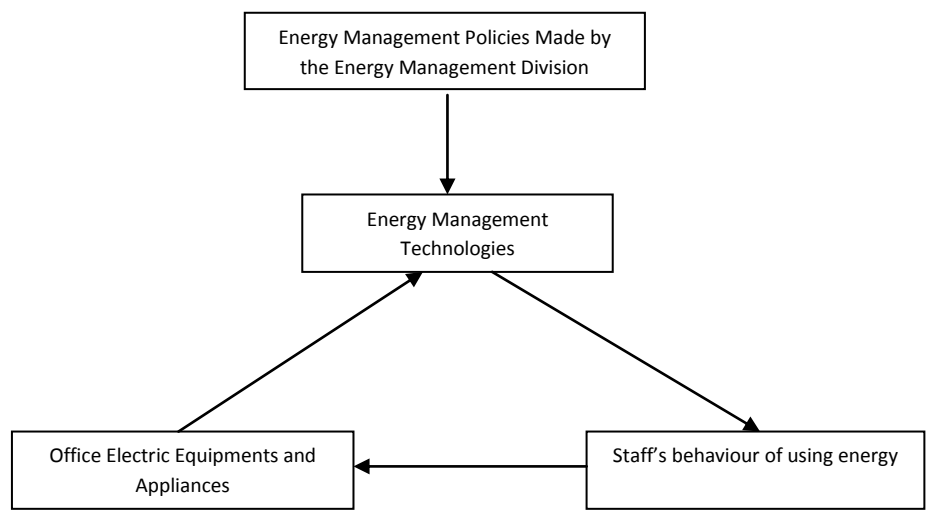

Figure 1: The Four Elements in Office Energy Consumption

## 2. Office Energy Consumption: A Mathematical Explanation

The energy consumption in an office is caused by the operation of various types of electric equipment and appliances (e.g. electric heaters, computing equipment and lights) in the office, which in turn control by the users' behaviour. Firth et al (2008) carry out a study on the types of home electric appliances and how these electric appliances contribute to the overall electricity consumption in a household. Firth et al (2008) classify home electric appliances into four categories based on their pattern of use: continuous appliances; standby appliances; cold appliances; and active appliances. *Continuous appliances* refer to electrical appliances such as clock, burglar alarms and broadband Internet modems which require constant amount of electricity. *Standby appliances* refer to electrical appliances such as televisions and game consoles which have three modes of operation: in use, on standby, or switched off; standby appliances consume electricity when they are in the modes of "in use" and "on standby", and some time even in the mode of "switched off" (e.g. Nintendo Wii game console); the only certain means to prevent them from consuming electricity is to disconnect their power supply. *Cold appliances* refer to electric appliances such as fridges and freezers which are continuously in use but do not consume constant amount of electricity; instead their electricity consumption cycles between zero and a preset level. Active appliances refer to electrical appliances such as lights, kettles and electrical cookers which are actively switched on or off by users and are clearly either in use or not in use; they do not have standby mode and when switched off they do not consume electricity at all. The electric equipment and appliances in office buildings (computing equipment, lights and security devices) have the same patterns of energy consumption as home electric appliances. Drawing on the idea of Firth et al (2008), we see the electricity consumed by continuous appliances (e.g. security cameras, information displays and computer servers) and cold appliances (e.g. refrigerators) as base consumption, because these kinds of electric equipment and appliances (we term them *base appliances*) have to be switched on all the time; and we see the electricity consumed by active appliances (e.g. lights) and standby appliances (e.g. desktop computers and printers) as flexible consumption, because these kinds of electric equipment and appliances (we term them *flexible appliances*) can be switched on/off at any time, depending on the behaviour of users. Thus, the total electricity consumption of an office in a certain period of time can be formulated as:

$$C_{\text{total}} = C_{\text{base}} + C_{\text{flexible}} \quad (1)$$

Where $C_{\text{base}}$ is the base electricity consumption and relates to the number and types of continuous and cold appliances the office has; $C_{\text{flexible}}$ is the flexible electricity consumption and relates to the number and types of active and standby appliances in the office.

Considering the individual active and standby appliances and the behaviour of their users, we can further break down the flexible consumption:

$$C_{\text{flexible}} = \beta_1 C_{\text{f1}} + \beta_2 C_{\text{f2}} + \beta_3 C_{\text{f3}} + \cdots + \beta_n C_{\text{fn}} \quad (2)$$

Where $C_{\text{f1}}, C_{\text{f2}}, C_{\text{f3}}, C_{\text{fn}}$ are the maximum electricity consumption of each flexible appliances; $n$ is the number of flexible appliances; and $\beta_1, \beta_2, \beta_3, \beta_n$ are the parameters reflecting the behaviour of the users. We range $\beta$ from 0 to 1. If $\beta$ is near 0, it means that the user of a flexible appliance always switches it off. If $\beta$ is close to 1, it means that the user of a flexible appliance always leaves it on.

Combining equation 1 and 2, we can derive a mathematical model to explain the electricity consumption in an office in a certain period of time:

$$C_{\text{total}} = C_{\text{base}} + (\beta_1 C_{\text{f1}} + \beta_2 C_{\text{f2}} + \beta_3 C_{\text{f3}} + \cdots + \beta_n C_{\text{fn}}) \quad (3)$$

Equation 3 explains how the behaviour of users can contribute to the overall electricity consumption in an office.

## 3. Agent-Based Simulation of Office Energy Consumption: A Case Study

As users' behaviour is significantly influenced by energy management technologies and energy management policies/regulations in an office building, the mathematical model above potentially integrates the four elements we mentioned before (i.e. energy management policies/regulations, energy management technologies, electric equipment/appliances and user behaviour) together and provide a theoretical basis for developing an agent-based simulation model of office energy consumption. Here we develop an agent-based model of office energy consumption based on the case of an academic building in the School of Computer Science, located in Jubilee Campus, the University of Nottingham. We choose this case simply because it is very convenient for us to carry out surveys to understand users' behaviour in the school, and also the Estate Office, who is responsible for the energy management in the University of Nottingham, kindly provides us with data about energy management technologies and really energy consumption in the school building.

## 3.1 Energy Consumption in the School Building

The School of Computer Science Building is situated in Jubilee Campus which was opened in 1999. Built on a previously industrial site, Jubilee Campus is an exemplar of brownfield regeneration and has impeccable green credentials. In terms of energy technologies, one important feature of the campus is the series of lakes which not only is the home of a variety of wildlife, but also provide storm water attenuation and cooling for the buildings in summers. Less visible, but equally important energy technologies are (1) roofs of the buildings covered by low-growing alpine plants which help insulate and maintain steady temperatures within the buildings throughout the year, (2) super-efficient mechanical ventilation systems, (3) lighting sensors to reduce energy consumption, and (4) photovoltaic cells integrated into the atrium roofs. Jubilee Campus has received many awards for its environment-friendly nature and energy efficiency of its buildings.

The School of Computer Science Building occupies a central position in Jubilee Campus, and is the academic home of some hundreds staff and students. The base energy consumption in the school building includes security devices, information displays, computer servers, shared printers and ventilation systems. The flexible energy consumption includes lights and office computers. In terms of energy management technologies, the school building is equipped with light sensors and half-hourly metering systems. Based on these energy management technologies, the energy management division (the Estate Office) has made automated lights energy management strategy (lights are automatically switched on when staff enter a room and switched off in 20 minutes after staff leave the room), and also select two environmental champions to promote energy saving awareness in the school.

## 3.2 Agent Based Model of Office Energy Consumption

We choose the first floor of the school of computer science as our case study. The building plan of the floor is as shown in Figure 2. One remark we should make here is that a large proportion of energy consumed in the school building is gas-fired heating, which is beyond our control. In this preliminary study, we focus on electricity consumption and ignore gas-fired heating. The details of the rooms and electric equipment and appliances are listed in Table 1.

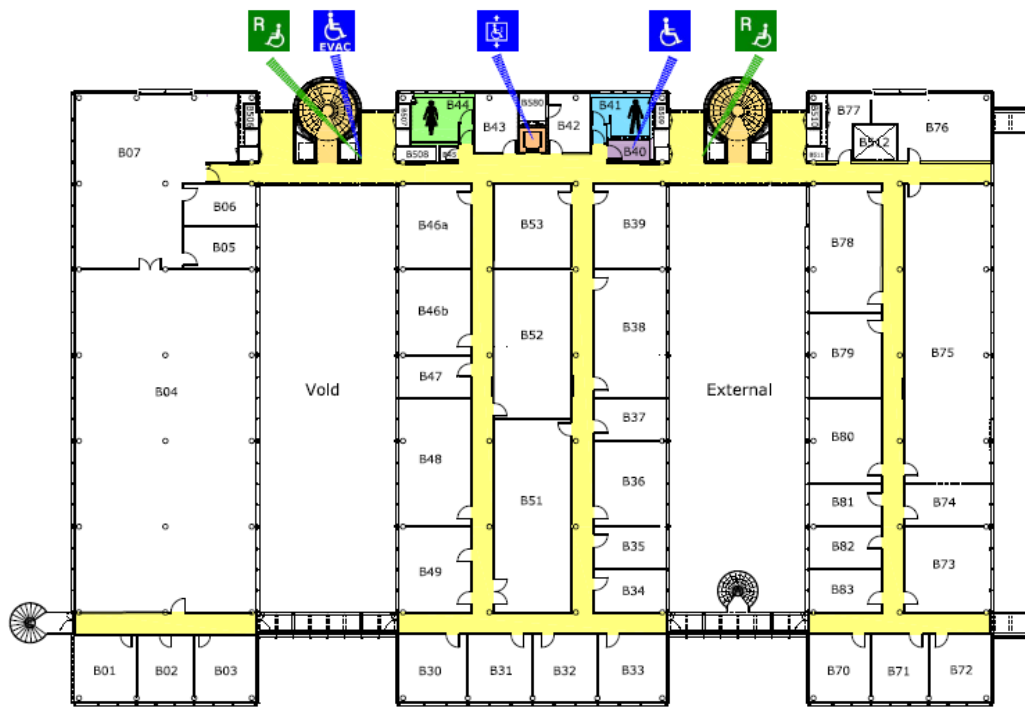

Figure 2: The Building Plan

Table 1: Details of Rooms and Electric Equipment and Appliances

| Item | Number |
| --- | --- |
| Rooms | 47 |
| Lights | 239 |
| Computers | 180 |
| Printers | 24 |
| Information Displays | 4 |
| Energy Users | 213 |

We design the model environment based on the office plan on the first floor of the school building (Figure 2), and implement the model in AnyLogic 6.5.0. The base electricity consumption of the school building is fixed, and the flexible consumption of the school building is caused by the interactions between flexible appliances (mainly lights and computers) and the energy users. We therefore focus on the flexible consumption, and design three types of agents: energy user agents, computer agents and light agents. These agents have been assigned to different rooms based on the office plan.

### 3.2.1 Behaviour of Energy User Agents

In order to understand the energy consumption behaviour of the energy users, we have carried out an extensive school wide empirical survey (questionnaire and observation). Our survey focuses on the energy use behaviour of the energy users (i.e. staff and PhD students) when they are in the School of Computer Science for work. Our empirical observation shows that each work day, energy users gradually enter the school building, walk through the corridors, and enter different offices for work. Their behaviour in different states can trigger the energy consumption of different electric appliances. Based on this empirical observation, we develop an energy user state chart to represent the behaviour of energy user agents, as shown in Figure 3.

We consider four different states of an energy user agent's behaviour in the model: out of school (*outOfSchool*), in corridor (*InCorridor*), in its own office (*InOwnOffice*) and in other rooms (*InOtherRooms*). In *outOfSchool* state, the energy user agent is not at work, thus does not trigger any energy consumption. In *InCorridor* state, as there are many lights in the corridors, the energy user agent's presence in the corridor triggers the lights on, which causes energy consumption. In *InOwnOffice* state, the energy user agent's presence in its own office triggers the office lights on, and its behaviour of using the computer in the office enables the computer in one of the following three modes: on, standby and off. Analogously, in *InOtherRooms* state the energy user agent's presence in other rooms (e.g. toilet, kitchen and lab) triggers the energy consumption of lights and computers (if any) in these rooms.

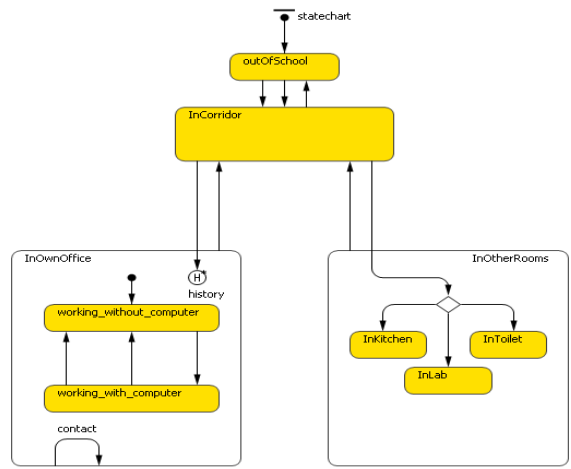

Figure 3: State Chart of Energy User Agent

The transitions between *outOfSchool* state and *InCorridor* state (both directions) is based on working timetable. Based on our empirical survey on working time, we have developed three stereotypes of energy user agents: early birds, timetable compliers, and flexible workers. Early birds (mainly cleaners, security staff and some hard working students and staff) often come to the school between 5 am and 9 am, and leave the school according to regular working time. Timetable compliers (mainly administrative staff and academic staff) often come to school between 9 am and 10 am, and leave the school building often at 5:30 pm. Flexible workers (mainly academic staff, research staff and PhD student) come to school at any time between 10 am and 1 pm, and leave the school at any time after their arrivals. Based on our statistics from our empirical survey, we assign relevant parameters for the energy user agents (Table 2). We also consider that each energy user agent has a very small probability ($p=0.02$) to work on Saturdays and Sundays. This consideration reflects the reality that a small number of hard working PhD student and research staff come to school on Saturdays and Sundays.

Table 2: Stereotypes of Energy User Agents and Parameters (I)

| Agent Stereotype | Percentage | Arrival Time | Leave Time |
| --- | --- | --- | --- |
| Early Birds | 8% | Monday to Friday, between 5am and 9am, random uniform distribution | Monday to Friday, between 5pm and 6pm, random uniform distribution |
| Timetable Compliers | 53% | Monday to Friday, between 9 am and 10 am, random uniform distribution | Monday to Friday, between 5pm and 6pm, random uniform distribution |
| Flexible Workers | 39% | Monday to Friday, between 10 am and 1 pm, random uniform distribution | Monday to Friday, between arrival time and 23pm, random uniform distribution |

The transition from *InCorridor* state to *InOwnOffice* state is the behaviour of entering own offices. In the simulation model we set the transition rule timeout = 2, which reflects the reality that normally after about 2 minutes walk in the corridors, an energy user can reach his/her own office. In *InOwnOffice* state, an energy user agent's presence can trigger the lights in its own office on. The energy user agent can either work with a computer (the substate *working_with_computer*), or work without a commuter (the substate *working_without_computer*). The transition from *working_without_computer* substate to *working_with_computer substate* is the behaviour of switching on the computer. We set the transition rule timeout = 2. This design is based on our empirical observation that normally an energy user switch on his/her computer within 2 minutes after he/she enters his/her office. The transitions from *working_with_computer* substate to *working_without_computer* substate is the behaviour of switching off or setting the computer on standby. For setting the computer on standby, the transition rule is a probability ($p = 0.05$) derived from our empirical survey. For switching off the computer, the transition rule is threshold control. We assume that each energy user agent has a personality parameter *energySavingAwareness*, ranging from 0 to 100, to represent its awareness on energy saving. If an energy user agent's *energySavingAwareness* is greater than the threshold, it has a large probability to switch off the computer when it does not need to use the computer. In the simulation, the threshold is an adjustable slide with value ranging from 0 to 100. Based on our empirical survey on staff's energy awareness, we create four stereotypes of energy user agent for the simulation model: Environmental Champion, Energy Saver, Regular User, and Big User. Different stereotypes of energy user agents have different levels of *energySavingAwareness*, and the probabilities for them to switch off the electric appliances that they do not have to use are different, as shown in Table 3. In *InOwnOffice* state, an energy user agent can also interact with other energy user agents. Our empirical survey shows that in terms of energy issues in the school, the most widely used interacting means is using emails in offices. We thus use an internal transition* "contact" within the *InOwnOffice* state to reflect the interactions between energy user agents. The larger *energySavingAwareness* an energy user agent has, the larger probability it will send email about energy saving issues to other energy user agents who have interactions with it. We set the interacting social network type as "small world". Based on the statistics from our empirical survey, we assign relevant parameters for these stereotypes of energy user agents (Table 3):

The transition from *InOwnOffice* state to *InCorridor* state reflects the behaviour of leaving own office. For an energy user agent, this can happen at any time between its arrival time and leave time. Thus the transition rule is a discrete event and the probability for its happening determined by its arrival time and leave time. Here we consider two kinds of leaves: temporally leave and long time leave. Temporally leave means that the energy user agent leaves its own office for less than 20 minutes, while long time leave means that it leaves its own office for more than 20 minutes. According to our empirical observation, people who temporally leave their offices do not usually switch off electric appliances, but people who leave their offices for a relatively long time do.

Table 3: Stereotypes of Energy User Agents and Parameters (II)

| Stereotype of Agent | Percentage | energySavingAwareness | Probability of Switching Off Unnecessary Electric Appliances | Probability of Sending Email about Energy Issues to Others |
|---|---|---|---|---|
| Environment Champion | 1% | Between 95 and 100, random uniform distribution | 0.95 | 0.9 |
| Energy Saver | 8% | Between 70 and 94, random uniform distribution | 0.7 | 0.6 |
| Regular User | 31% | Between 30 and 69, random uniform distribution | 0.4 | 0.2 |
| Big User | 60% | Between 0 and 29, random uniform distribution | 0.2 | 0.05 |

The transition from *InCorridor* state to *InOtherRooms* state reflects the behaviour of using other facilities such as kitchens, toilets and labs in the school. For an energy user agent, this behaviour can also happen at any time between its arrival time and leave time. Again we consider the transition rule as a discrete event and the probability for its happening determined by its arrival time and leave time.

The transition from *InOtherRoom* state to *InCorridor* state reflects the behaviour of stopping using other facilities and leaving the facility rooms. The transition rule is timeout. Here we set the time range from 1 to 10 (random uniform distribution), which reflect the reality that a user usually finish using facilities such as toilets or kitchens within 10 minutes.

Basically the state chart, which we have developed based on our empirical survey, can well reflects the all the behaviour of a real energy user with regards to energy consumption when he/she works in the school building.

### 3.2.2 Behaviour of Light Agents

In the simulation model, we treat the lights in the school building as passive agents, which means that these agents do not have proactive behaviour. Instead, their behaviour is passive reactions to the behaviour of energy user agents. Their behaviour is relatively simple, as they are just passively off and on, as shown in the light state chart (Figure 4)

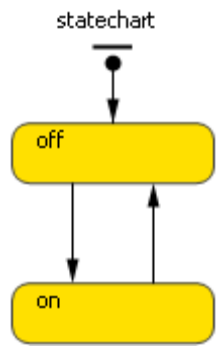

Figure 4: State Chart of Light Agents

For a light, the transition from *off* state to *on* state is associated with the presence of energy user agents if the light is automated by light censors, or with the behaviour of switching on if there is a light switch which enables energy user agents to have control over the light. Conversely, the transition from *on* state to *off* state is associated with the leave of energy user agents if the light is automated by light censors, or with the behaviour of switching off if there is a light switch. According to the data provided by the Estate Office, when a light is on its power is 60 Watts; when it is off its power is 0.

### 3.2.3 Behaviour of Computer Agents

Similar to lights, computers have also been treated as passive agents in the simulation model. Computers can be switched on, off, and be set standby. Thus in the state chart of a computer, we consider these three states, as shown in Figure 5.

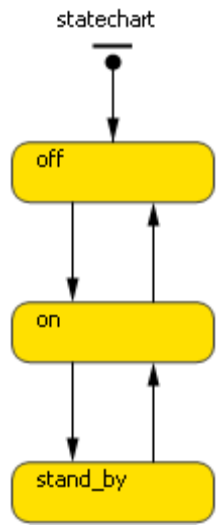

Figure 5: State Chart of Computer Agents

For a computer, the transitions between *off*, *on* and *stand_by* states are related to energy user agents' behaviour of using the computer. According to the data provided by the Estate Office, when a computer is off, its overall power is 0; when it is on standby mode, its overall power is 25 Watts; and when it is on, its overall power is 70 Watts.

### 3.2.4 Model Implementation

The model has been implemented in the simulation package AnyLogic 6.5.0 on a standard PC with Windows XP SP3. We set each time step in the simulation model as one minute, and simulate the daily work of staff in the school of computer science and observe and analyse how their behaviour can result in the system level electricity consumption of the whole school. The light agents and computer agents are assigned to each room, based on their real physical distribution in the school. The energy user agents come to school every morning, walk through the corridors and enter their own offices for work. They may also leave their offices, walk through the corridors and enter other rooms for using facilities such as toilet and kitchens. They interact with each other in terms of energy issues in the school, and this kind of interaction can change their *energySavingAwareness*. They also interact with passive light and computer agents, and this kind of interaction can directly result in the system level electricity consumption of the school. An overview of the model is shown in Figure 6. We also animate the energy user agents, and the interface of the model is shown in Figure 7.

## 4. Simulation Experiments

With the model, we have carried out three sets of experiments. We use these sets of experiments to test the validity of the model, design energy management strategies for the Estate Office and help the Estate Office gain insights about the energy consumption in the school.

**Experiment 1: Reproduce the current energy management strategy of the school**

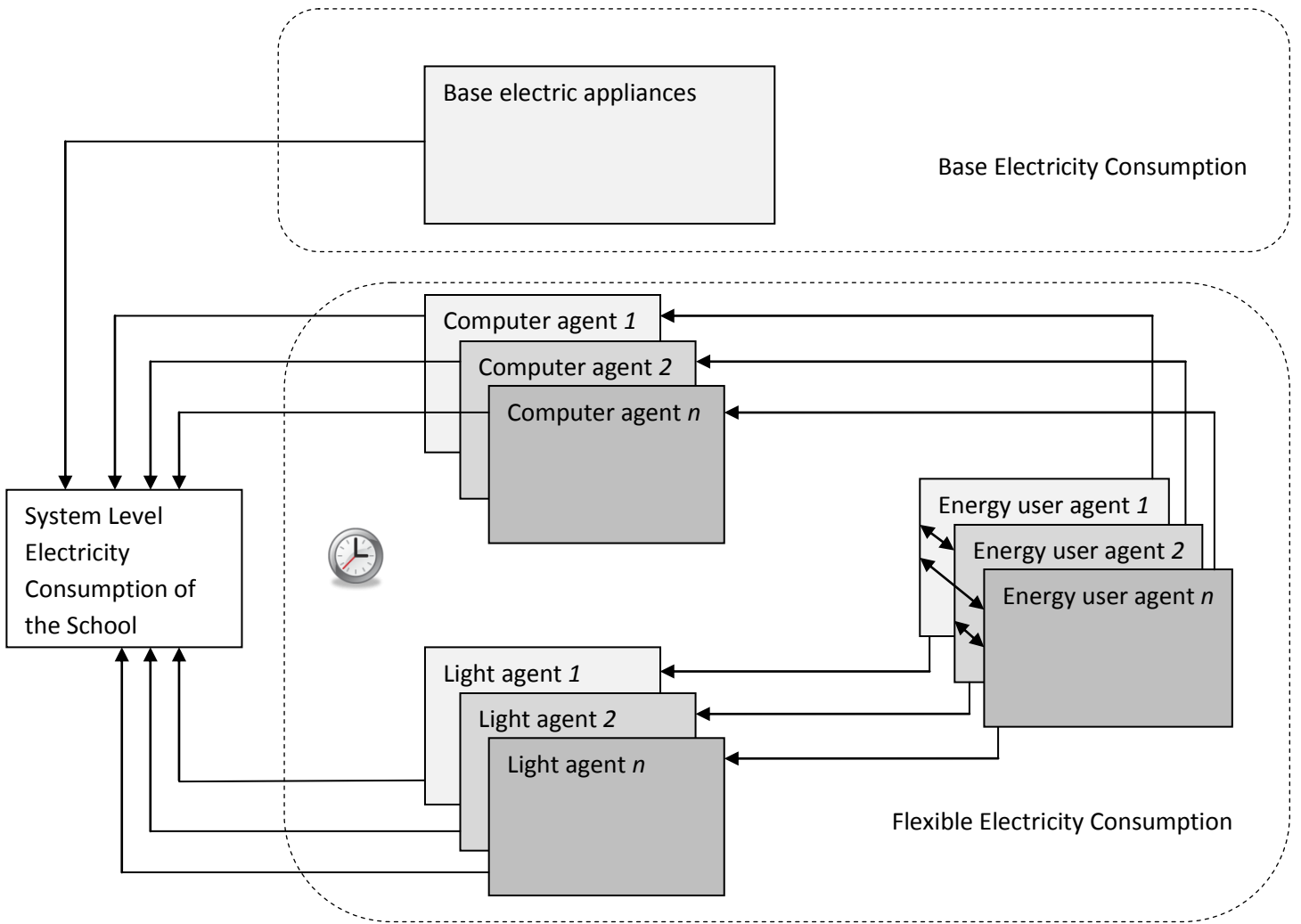

Figure 6: Overview of the Model

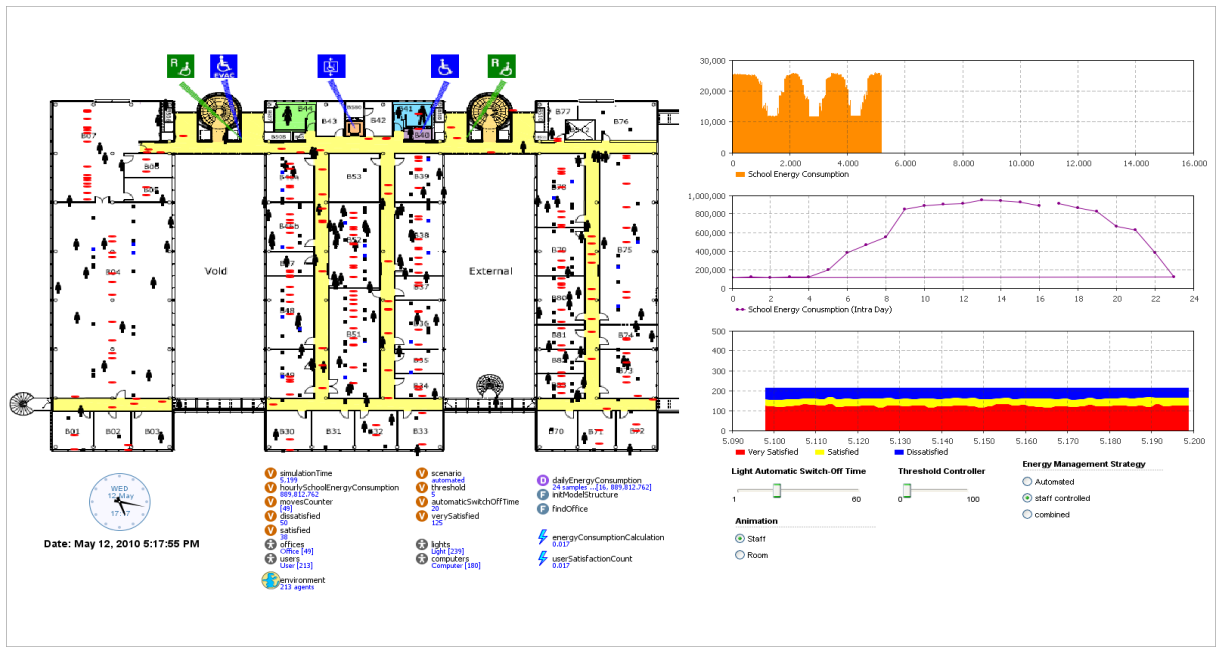

Figure 7: Interface of the Model

Currently, the computer science school has equipped with light sensors which automatically switch on the lights when they detect the presence of staff, and switch off the lights when they detect the absence of staff for 20 minutes. Thus based on the light sensor technology, the Estate Office has adopted the *automated* energy management strategy in the school of computer science. In that sense, staff do not have control over the switch-on/off of the lights, and they only have control over the switch-on/off the computers. Our first set of experiments focuses on this and aims to reproduce the energy management strategy. We set the model in the "automated" scenario, run the model and plot the system level school electricity consumption in Figure 8, from which we can see that the pattern of simulated school electricity consumption is quite similar to the real school electricity consumption provided by the Estate Office (Figure 8). The similarity from the comparison signifies that we have successfully reproduced the current energy management strategy in the school of computer science, and also validates our model.

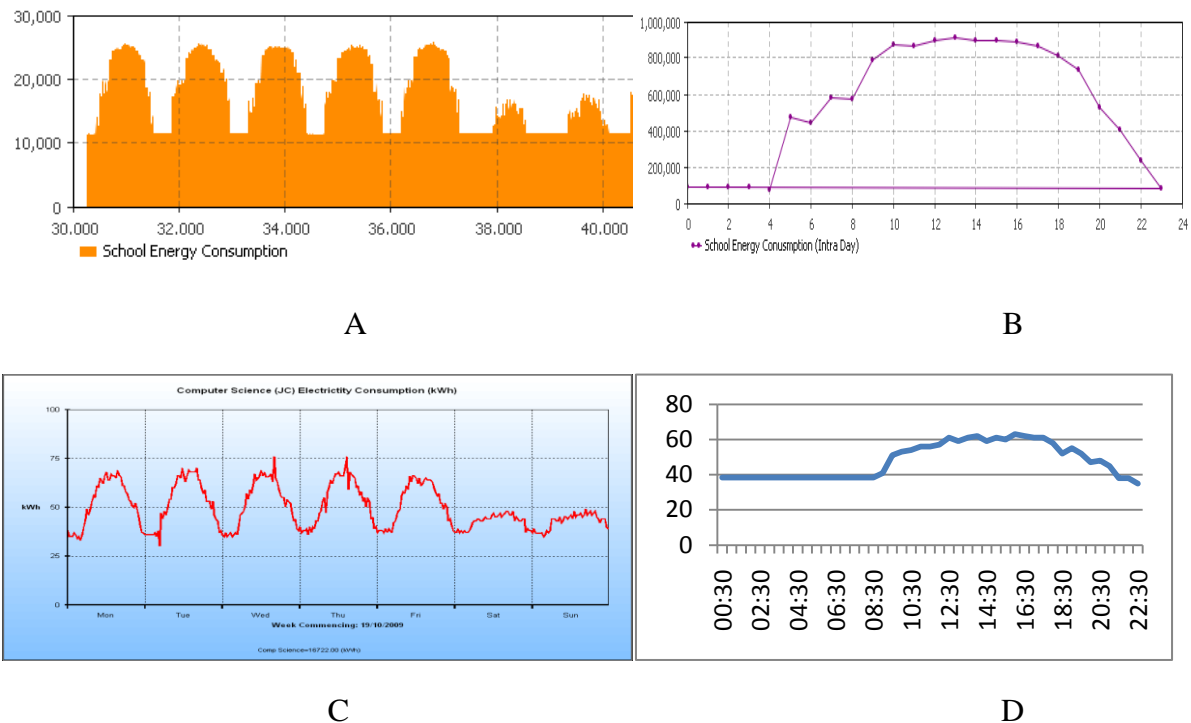

Figure 8: Comparison of Simulation Results and Empirical Results (Experiment 1)

Note: in Figure 8, A is the simulation result of overall electricity consumption in one week (Monday to Sunday), C is the real electricity consumption in one week (provided by the Estate Office); B is the simulation result of overall electricity consumption in a day, and D is the real electricity consumption in a day (provided by the Estate Office)

**Experiment 2: Automated Strategy vs. Staff-Controlled Strategy**

A significant body of research (e.g. Gritchely, et al, 2006; Linden, et al, 2006; Wilhit, et al, 1996; Peeters, et al, 2008; Boait and Rylatt, 2010) shows that energy users have difficulty with manual energy control and management, and thus argues for an automated energy management strategy which maintains efficient energy consumption to satisfactorily meet energy users' energy needs and meanwhile minimizes the cost of energy without any user intervention in both homes and office buildings. However, these studies mentioned before focus on domestic heating systems which traditionally have sophisticated control units such as programmers and temperature sensors. In terms of energy management strategies for lighting, there is a lack of sound evidence that manual controlled strategy is less efficient than automated strategy. There seems to be a debate between the automated strategy and staff-controlled strategy. On the one hand, anecdotally, many people, particularly technical people from the Estate Office, strongly believe that automated lighting is more energy-efficient than staff-controlled lighting. On the other hand, our empirical survey shows that some people in the school of computer science believe that if they can control the switch-on/off of the lights, the electricity consumption in the school would be less, as under the automated lighting strategy the lights are off only after 20 minutes of their leave, which causes unnecessary consumption of electricity. Our second set of experiments focuses on the debate: we compare the simulation results under the two different lighting management strategies. The rationale for the two strategy scenarios is as follows: in the automated lighting management strategy scenarios, lights in an office are off 20 minutes after the last occupying energy user agent' leave, while in the staff-controlled lighting management strategy, lights in an office are off immediately according to a probability after the leave of the last occupying energy user agent. The probability of is determined by the *energySavingAwareness* of the last occupying energy user agent. The larger *energySavingAwareness* the last occupying energy user agent has, the larger the probability is. The probabilities are assigned based on Table 3. This design reflects the reality that under the staff-controlled lighting management strategy, staff can switch off the lights when they leave their offices. The more the staff are concerned about energy saving, the more possible they will switch off the lights when they leave their offices. The comparison of the simulation results of the two scenarios are shown in Figure 9, from which we can see that although the peak time energy consumption is almost the same, the energy consumption in staff-controlled lighting management strategy scenario is substantially higher than that in automated lighting management strategy scenario. Thus, one energy management strategy implication we can draw from the simulation is that in the current circumstance, the automated lighting management strategy is more energy-efficient than staff-controlled energy management strategy.

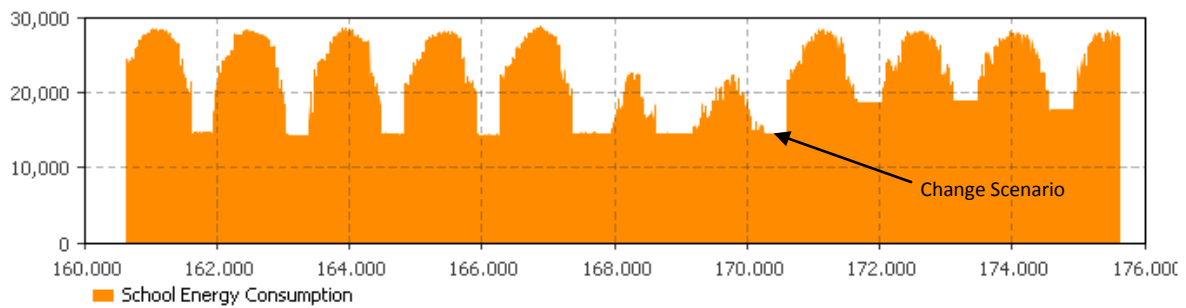

Figure 9: Simulation Results (Experiment 2A)

Note: In the simulation, we have changed the scenario from "automated" to "staff-controlled". We can see from the simulation result that clearly the electricity consumption in "staff-controlled" scenario is substantially increased, although the peak time energy consumption is almost the same

We note that the probabilities for staff to switch off lights when they leave their offices are related to their *energySavingAwareness.* One question to which we would like to seek answer from the model is that: if we increase the energy user agents' *energySavingAwareness* by enhance the interactions about energy issues between energy user agents, is automated lighting management strategy still more energy-efficient than staff-controlled lighting management strategy? We increase the contact rate (i.e. the frequency of contact in a certain simulation period), run the model and gain the simulation results in Figure 10, which shows a negative answer to that question. Increasing energy users agents' *energySavingAwareness* through social interactions can significantly reduce the overall energy consumption of the school. The senior management of the university has already realized the importance of increasing staff's energy saving awareness, thus a university-wide campaign called "gogreener" has been carried out and two environmental champions have been appointed in each school to monitor the energy consumption of the school and enhance the interactions of staff in terms of energy issues.

**Experiment 3: Understand the proportions of electricity consumed by lights and computers**

The Estate Office has installed some half-hourly electricity meters in the school building to monitor the electricity consumption in the school of computer science. Although these meters can tell us how much electricity consumed by the school, they are not able to tell us how much electricity consumed by computers and how much electricity consumed by lights, which is also a question the Estate Office keen to know. As indicated in the model, the electricity consumed by lights and computers is related to behaviour of energy user agents, which makes it hard to be measured in a simply way. With the help of the simulation model, we can gain some insights into this issue. We run the model in *automated* scenario (i.e. the current lighting management strategy of the school), and plot the electricity consumption by both lights and computers in Figure 11, from which we can see the proportions of electricity consumed by computers and lights vary over time. We also plot one week electricity consumption, as shown in Figure 11.

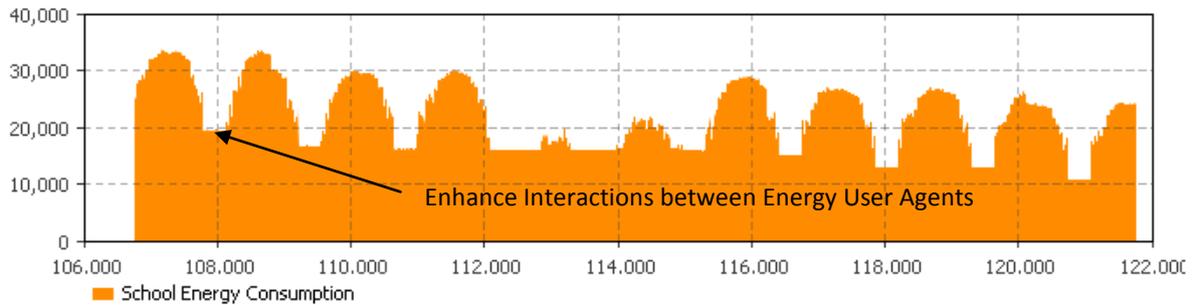

Figure 10: Simulation Results (Experiment 2B)

Note: In the simulation, we have increased the "contact rate" so as to enhance the interactions between energy user agents. We can see from the simulation result that obviously the overall electricity consumption is decreasing.

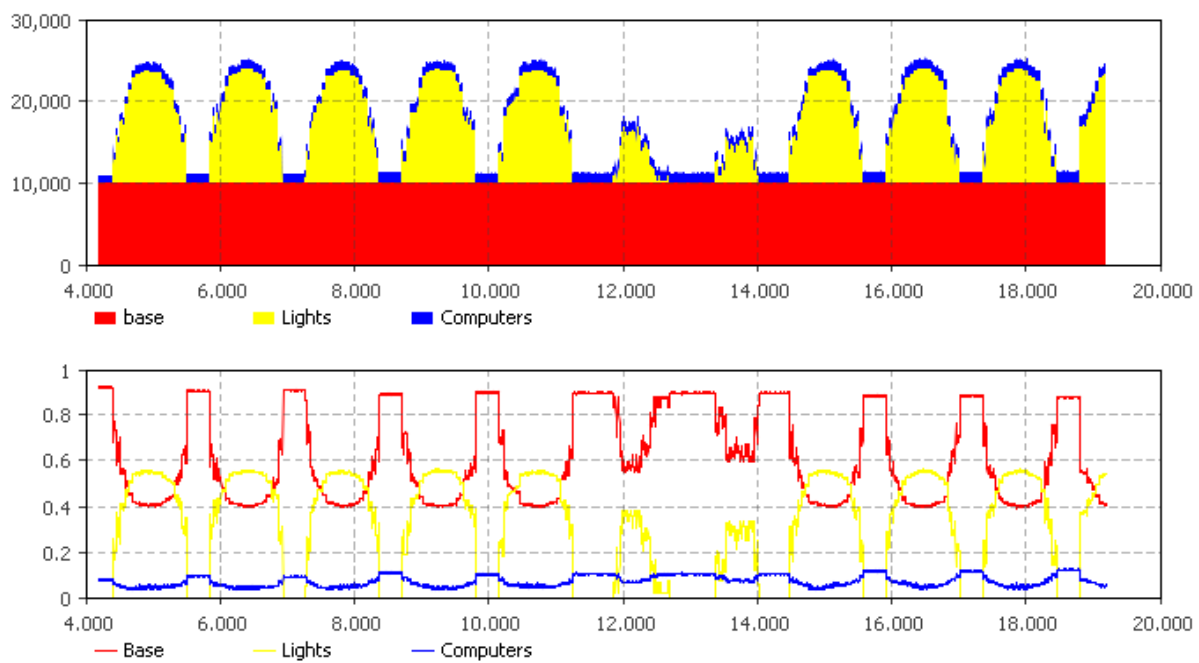

Figure 11: Electricity Consumed by Computers, Lights and Based Electric Appliances

Note: The figure on the top is the amounts of electricity consumed by lights, computers and based electric appliances, while the figure on the bottom is the percentages. From the simulation results we can see that in the evenings and weekends, most of the electricity consumption is base consumption (92%); in the daytime (Monday to Friday), the electricity consumed by computers (7%) is much less than that consumed by lights (55%).

## 5. Discussion

**Theory Based Agents vs. Empirical Survey Based Agents**

In social simulation, many agent-based simulation studies develop the agents in their models based on well-established social theories. For example, in marketing researchers develop consumer agents based on social psychological theories such as social comparison, imitation (e.g. Janssen and Jager, 1999, 2002, 2003) and the theory of planned behaviour (e.g. Zhang and Zhang, 2007; Zhang & Nutall, 2008). In energy economics, Bunn and Oliveira (2001) develop electricity market agents (i.e. electricity generating companies and electricity suppliers) based on market bidding theory to simulate the New Electricity Trading Arrangements (NETA) of England and Wales. Clearly, the existence of these well-established social theories significantly facilitates the development of the social agents. In this particular case of modelling office energy consumption, however, there are no well-established theories about staff's behaviour of using energy for us to draw on. We thus have to conduct time-consuming and costly empirical surveys and observations on the behaviour of real world objects (i.e. staff and PhD students in the school) and develop agents based on our empirical survey. Our empirical survey covered most of the aspects of staff's energy use in the school. As a result, the state chart we have developed to represent the behaviour of energy user agents in our model has a very strong empirical basis. That is also the reason why our simulation results are quite similar to the real world observation. Empirical survey based agents are increasingly used in social simulation, particularly in operations management (e.g. Siebers, et al, 2009, 2010). Compared to traditional theory based agents, empirical survey based agents, although require much more time and work for the development, are easier to be calibrated and validated.

**Limitations**

Although we have developed the state chart to represent the behaviour of energy user agents based on a comprehensive empirical survey, clearly the agents in computer cannot perfectly replicate the real-life of energy users in the school. Therefore it is important to acknowledge the limitations of the model. Firstly, an energy user agent's stereotype in the model is fixed. In other words, in the simulation there is no way for an energy user agent to switch its stereotype (e.g. from an early bird to a flexible worker). But we note in the real world the switch of stereotypes can happen, although the probability for its happening is low. A second limitation is our assumption that enhancing interactions about energy issues between staff can increase staff's energy saving awareness. This assumption is true in the situation where energy users have to bear the cost of energy, as some research on energy efficiency in domestic sector has already proved it. However, while working in office buildings staff do not have to bear the cost of energy, which may result in the assumption in question. Currently we have not found any sound evidence to support this assumption.

**Further Research**

The agent-based model of office energy consumption described in this paper has potency for further development. Theoretically, we can incorporate more flexible electric appliances and more complex human-electric appliance interactions into the model, which will make the model more applicable, and can be applied to modelling a large organisation which has very complex behaviour of consuming energy, e.g. large number of staff and complex energy management strategies/regulations. We can also extend the model to study gas-fired heating systems in office building by drawing an analogy between electricity consumption and gas consumption. Moreover, we can add more psychological factors into the energy user agents, and study how to optimize energy consumption for an organisation while maintain its staff's satisfaction about energy use.

## 6. Conclusion

This paper has described an agent-based model of office energy consumption. We began the paper with an argument for an integration of the four elements involved in office energy consumption, and then presented a mathematical model to explain the energy consumption in office building. We developed an agent-based model of office energy consumption based on the case of School of Computer Science, in Jubilee Campus, the University of Nottingham, and presented the simulation results. Along the way, we focused on two objectives. One is the integration of the four elements involved in office energy consumption with an agent-based model. The other one is utilizing the developed model to study practical energy management for an organisation. From the research we reported in this paper, we conclude that, although it is not possible to perfectly replicate the real organisation, agent-based simulation as a novel approach which integrates the four elements involved in office energy consumption, is a very useful tool for office building energy management.